\documentclass[runningheads]{llncs}
\usepackage{subcaption}
\usepackage{graphicx}

\usepackage{listings}

\usepackage{array,booktabs}
\newcolumntype{M}[1]{>{\centering\arraybackslash}m{#1}} 
\usepackage{subcaption} 

\usepackage{afterpage}

\usepackage[normalem]{ulem}
\useunder{\uline}{\ul}{}

\usepackage{comment}
\usepackage{amsmath,amssymb} 
\usepackage{color}

\usepackage{pifont}
\newcommand{\cmark}{\ding{51}}%
\usepackage{csquotes}
\usepackage{tikz}
\usetikzlibrary{arrows.meta,
                positioning,
                shapes}

\usepackage[ruled,linesnumbered]{algorithm2e}

\usepackage[accsupp]{axessibility}  

\usepackage{url}

\usepackage{hyperref} 

\newcommand\nnfootnote[1]{  
  \begin{NoHyper}
  \renewcommand\thefootnote{}\footnote{#1}%
  \addtocounter{footnote}{-1}%
  \end{NoHyper}
}

\begin{document}
\pagestyle{headings}
\mainmatter
\def\ECCVSubNumber{1}  

\title{One Ontology to Rule Them All:\\
Corner Case Scenarios for Autonomous Driving} 

\titlerunning{One Ontology To Rule Them All}
%
\author{Daniel Bogdoll\inst{1,2}\textsuperscript{\textasteriskcentered} \and
Stefani Guneshka\inst{2}\textsuperscript{\textasteriskcentered} \and
J. Marius Zöllner\inst{1,2}}
\authorrunning{Bogdoll et al.}
%
\institute{FZI Research Center for Information Technology, Germany\\
\email{bogdoll@fzi.de}\\
\and
Karlsruhe Institute of Technology, Germany}
\maketitle

\nnfootnote{\textasteriskcentered~These authors contributed equally}

\begin{abstract}
The core obstacle towards a large-scale deployment of autonomous vehicles currently lies in the long tail of rare events. These are extremely challenging since they do not occur often in the utilized training data for deep neural networks. To tackle this problem, we propose the generation of additional synthetic training data, covering a wide variety of corner case scenarios. As ontologies can represent human expert knowledge while enabling computational processing, we use them to describe scenarios. Our proposed master ontology is capable to model scenarios from all common corner case categories found in the literature. From this one master ontology, arbitrary scenario-describing ontologies can be derived. In an automated fashion, these can be converted into the OpenSCENARIO format and subsequently executed in simulation. This way, also challenging test and evaluation scenarios can be generated.
\end{abstract}

\section{Introduction}
\label{sec:introduction}
For selected Operational Design Domains (ODD), autonomous vehicles of the SAE level~4~\cite{SAE_J3016_2021_Standard} can already be seen on the roads~\cite{Waymo_One_2022_Web}. However, it remains highly debated, how the safety of these vehicles can be shown and steadily improved in a structured way. In Germany, the first country with a federal law for level 4 autonomous driving, the safety of such vehicles needs to be demonstrated based on a catalog of test scenarios~\cite{bundesregierung_entwurf_2021,bundesrat_verordnung_2022}. However, the coverage of rare, but highly relevant corner cases~\cite{Karpathy_Tesla_2019_Web} in scenario-based descriptions poses a significant challenge~\cite{pretschnerTestsFurAutomatisierte2021}. Data-driven, learned scenario generation approaches currently tend to focus on adversarial scenarios with a high risk of collision~\cite{multimodal_Ding_2021,Wang_2021_CVPR,geiger_eccv_king,Rempe_2022_CVPR}, neglecting other forms of corner cases. While there exist comprehensive taxonomies on the types and categories of corner cases~\cite{breitenstein_corner_2021,Heidecker_2021}, there exist no generation method tailored to these most important long-tail scenes and scenarios. Based on this, the verification and validation during testing and ultimately the scalability of autonomous driving systems to larger ODDs in real world deployments remain enormous challenges. To tackle these challenges, it is necessary to generate a large variety of rare corner case scenarios for the purposes of training, testing, and evaluation of autonomous vehicle systems. As shown by Tuncali et al.~\cite{8911483}, model-, data-, and scenario-based methods can be used for this purpose. An extensive overview on these can be found in~\cite{Bogdoll_Description_2021_ICCV}. However, the authors find that, "while there are knowledge-based descriptions and taxonomies for corner cases, there is little research on machine-interpretable descriptions"~\cite{Bogdoll_Description_2021_ICCV}.

To fill this gap between knowledge- and data-driven approaches for the description and generation of corner case scenarios\footnote{We follow the definitions of scene and scenario by Ulbrich et al.~\cite{Ulbrich:2015d}, where a scene is a snapshot, and a scenario consists of successive scenes}, we propose the first scenario generation method which is capable of generating all corner case categories, also called levels, described by Breitenstein et al.~\cite{breitenstein_corner_2021} in a scalable fashion, where all types of scenarios can be derived from a single master ontology. Based on the resulting scenario-describing ontologies, synthetic data of rare corner case scenarios can be generated automatically. This newly created training data hopefully contributes to an increased robustness of deep neural networks to anomalies, helping make deep neural networks safer. For a general introduction to ontologies in the domain of information sciences, we refer to \cite{onto101}. More details can be found in~\cite{Guneshka_Ontology_2022_BA}. All code and data is available on \href{https://github.com/fzi-forschungszentrum-informatik/corner_case_ontology}{GitHub}. 

\vspace{3mm} 
The remainder of this work is structured as follows: In Section~\ref{sec:related_work}, we provide an overview of related ontology-based scene and scenario description methods and outline the identified research gap. In Section~\ref{sec:method}, we describe how our proposed master ontology is designed and the automated process to design and generate scenario ontologies from it. In Section~\ref{sec:evaluation}, we demonstrate how different, concrete scenarios can be derived from the master ontology and how the resulting ontologies can be used to execute these in simulation. Finally, in Section~\ref{sec:conclusion}, we provide a brief summary and outline next steps and future directions.

\section{Related Work}
\label{sec:related_work}
While there exist many ways of describing scenarios~\cite{Bogdoll_Description_2021_ICCV}, ontologies are the most powerful way of doing so, as these are not only human- and machine-readable, but also extremely scalable for the generation of scenarios, when used in the right fashion~\cite{Hermann_Using_2022_DATE}. Ontologies are being widely used for the description of scenarios. In the work of Bagschik et al.~\cite{SceneCreation}, an ontology is presented which describes simple highway scenarios based on a set of pre-defined keywords. In a later work, Menzel et al.~\cite{functional_to_logical} extend the concept to generate OpenSCENARIO and OpenDRIVE scenarios, while many of the relevant details were not modelled in the ontology itself, but in post-processing steps. For the description of the surrounding environment of a vehicle, Fuchs et al.~\cite{fuchs_ontology} especially focus on lanes and occupying traffic participants, while neglecting their actions. Li et al.~\cite{Li_TestGeneration_2020} also create scenarios which are executed in a simulation environment, covering primarily situations, where sudden braking maneuvers are necessary. Thus, their ontology is very domain-specific. They build upon their previous works~\cite{Tao_Industrial_2019,Wotawa2018FromOT,Klueck_Using_2018}. Tahir and Alexander~\cite{Zaid_Intersection_2022} propose an ontology that focuses on intersections due to their high collision rates. They show that their scenarios can be executed in simulation, while focusing on changing weather conditions. While they claim to have developed an ontology, the released code~\cite{zahid_repo} only contains scripted scenarios, which might be derived from an ontology structurally. Hermann et al.~\cite{Hermann_Using_2022_DATE} propose an ontology for dataset creation, with a demonstrated focus on pedestrian detection, including pedestrian occlusions. Their ontology is structurally inspired by the Pegasus model~\cite{pegasus} and consists of 22 sub-ontologies. It is capable of describing a wide variety of scenarios and translate them into simulation. However, since the ontology itself is neither described in detail nor publicly available, it does not become clear whether each frame requires a separate ontology or whether the ontology itself is able to describe temporal scenarios. In the OpenXOntology project by ASAM~\cite{asam_openxontology}, an ontology is being developed with the purpose to unify their different products, such as OpenSCENARIO or OpenDRIVE. Based on the large body of previous work in the field of scenario descriptions, this ontology is promising for further developments. However, at the moment, it serves the purpose of a taxonomy. Finally, Gelder et al.~\cite{Gelder_Towards_2022} propose an extensive framework for the development of a "full ontology of scenarios". At the moment, they have not developed the ontology itself yet, which is why their work cannot be compared to existing ontologies.

\begin{table}[h!]
\centering
\resizebox{\columnwidth}{!}{%
\begin{tabular}{@{}llcccccc@{}}
\toprule
Authors             & Year & \begin{tabular}[c]{@{}c@{}}Temporal Scenario\\ Description\end{tabular} & \begin{tabular}[c]{@{}c@{}}Arbitrary\\ Environments\end{tabular} & \begin{tabular}[c]{@{}c@{}}Arbitrary\\ Objects\end{tabular} & \begin{tabular}[c]{@{}c@{}}Scenario\\ Simulation\end{tabular} & \begin{tabular}[c]{@{}c@{}}Corner Case\\ Categorization\end{tabular} & \begin{tabular}[c]{@{}c@{}}Ontology\\ available\end{tabular} \\ \midrule
Fuchs et al.~\cite{fuchs_ontology}        & 2008 & \textbf{-}                                                              & \textbf{-}                                                       & \textbf{\cmark}                              & \textbf{-}                                                    & \textbf{-}                                                           & \textbf{-}                                                \\
Hummel~\cite{hummel_intersections}              & 2010 & \textbf{-}                                                              & \textbf{\cmark}                                   & \textbf{-}                                                           & \textbf{-}                                                             & \textbf{-}                                                           & \textbf{-}                                                \\
Hülsen et al.~\cite{traffic_ontology}       & 2011 & \textbf{\cmark}                                          & \textbf{\cmark}                                   & \textbf{-}                                                           & \textbf{-}                                                    & \textbf{-}                                                           & \textbf{-}                                                \\
Armand et al.~\cite{ontology_based_context_awareness}       & 2014 & \textbf{-}                                                                       & \textbf{-}                                                       & \textbf{-}                                                  & \textbf{-}                                                    & \textbf{-}                                                           & \textbf{-}                                                \\
Zhao et al.~\cite{core_ontologies}         & 2017 & \textbf{-}                                                              & \textbf{\cmark}                                   & \textbf{-}                                                           & \textbf{-}                                                             & \textbf{-}                                                           & \textbf{\cmark}                                                \\
Bagschik et al.~\cite{SceneCreation}     & 2018 & \textbf{\cmark}                                          & \textbf{-}                                                       & \textbf{-}                                                  & \textbf{-}                                                    & \textbf{-}                                                           & \textbf{-}                                                \\
Chen and Kloul~\cite{highway_ontology}       & 2018 & \textbf{\cmark}                                          & \textbf{-}                                                       & \textbf{-}                                                           & \textbf{-}                                                             & \textbf{-}                                                           & \textbf{-}                                                \\
Huang et al.~\cite{huang_ontology}        & 2019 & \textbf{\cmark}                                                           & \textbf{-}                                                    & \textbf{-}                                               & \textbf{-}                                                 & \textbf{-}                                                           & \textbf{-}                                                \\
Menzel et al.~\cite{functional_to_logical}       & 2019 & \textbf{\cmark}                                          & \textbf{\cmark}                                   & \textbf{-}                                                  & \textbf{\cmark}                                & \textbf{-}                                                           & \textbf{-}                                                \\
Li et al.~\cite{Li_TestGeneration_2020}             & 2020 & \textbf{\cmark}                                          & \textbf{\cmark}                                   & \textbf{-}                                                           & \textbf{\cmark}                                & \textbf{-}                                                           & \textbf{-}                                                \\
Tahir and Alexander~\cite{Zaid_Intersection_2022} & 2022 & \textbf{\cmark}                                          & \textbf{-}                                                                & \textbf{-}                                                           & \textbf{\cmark}                                & \textbf{-}                                                           & \textbf{-}                                                \\
Hermann et al.~\cite{Hermann_Using_2022_DATE}      & 2022 & \textbf{-}                                                           & \textbf{\cmark}                                                    & \textbf{\cmark}                              & \textbf{\cmark}                                                 & \textbf{-}                                                           & \textbf{-}                                                   \\
ASAM~\cite{asam_openxontology}                & 2022 & \textbf{-}                                                              & \textbf{-}                                                       & \textbf{-}                                                  & \textbf{-}                                                    & \textbf{-}                                                           & \textbf{-}                                                   \\ \midrule
Proposed Ontology   &      & \textbf{\cmark}                                          & \textbf{\cmark}                                   & \textbf{\cmark}                              & \textbf{\cmark}                                & \textbf{\cmark}                                       & \textbf{\cmark}                               \\ \bottomrule
\end{tabular}%
}
\vspace{3mm} 
\caption{Comparison of related ontologies and our proposed ontology for scenario descriptions in the field of autonomous driving.}
\label{tab:sota-table}
\end{table}

Next to ontologies which are explicitly designed to describe scenarios, more exist which also focus on decision-making aspects. In this category, Hummel~\cite{hummel_intersections} developed an ontology capable of describing intersections to a degree, where the ontology can also be used to infer knowledge about the scenes. While this is a general attribute of ontologies, she provides a set of rules for the analysis. Hülsen et al.~\cite{traffic_ontology} also describe intersections based on an ontology, focusing on the road layout, while interactions between entities cannot be modeled in detail. In~\cite{ontology_based_context_awareness}, this issue is addressed, as Armand et al. focus on such interactions. They also propose rules to infer knowledge from their ontology. These rules are partly attributed to the decision-making of an ego vehicle, e.g., whether it should stop or continue. Due to their strong focus on actions and interactions, they struggle to describe complex scenarios in a more general way. Zhao et al.~\cite{core_ontologies} developed a set of three ontologies, namely Map, Car, and Control. Based on these, they are capable of describing complex scenes for vehicles only. While the scenes do contain temporal information, such as paths for vehicles, these are only broad descriptions and not detailed enough to model complex scenarios. Huang et al.~\cite{huang_ontology} present a similar work that is able to describe a wide variety of scenarios based on classes for road networks for highway and urban scenarios, the ego vehicle and its behavior, static and dynamic objects, as well as scenario types. However, it is designed to derive driving decisions from the descriptions instead of simulating these scenarios. Chen and Kloul~\cite{highway_ontology} on the other hand propose an ontology that is primarily designed to describe highway scenarios, with a special focus on weather circumstances. 

\vspace{3mm} 
To model corner case scenarios, the requirements for an ontology are very complex. In general, it needs to be able to describe all types of scenes and scenarios. For the temporal context, an ontology needs to be able to \textit{a) describe scenarios}. Furthermore, it needs to be able to \textit{b) describe arbitrary environments} and \textit{c) arbitrary objects}. Following an open world assumption, we define "arbitrary", in respect to environments and objects, as the possibility to include such without changing any classes or properties of the ontology. This means, e.g., referencing to external sources, such as OpenDRIVE files for environments or CAD files for objects. An ontology needs to be designed in a way that \textit{d) the described scenarios can also be simulated}. Finally, \textit{e) information about the corner case levels} needs to be included for details and knowledge extraction. In Table~\ref{tab:sota-table}, we provide an overview of the previously introduced ontologies related to these attributes and also mention, whether the ontology itself is published online. While some authors, such as~\cite{fuchs_ontology,hummel_intersections}, released their ontologies previously, the provided links do not contain them anymore, which is why we excluded outdated sources. A trend can be observed, where recent approaches focus more on the aspect of scenario simulation. However, to the best of our knowledge, there exists no ontology to date that is able to describe and simulate long-tail corner case events. Our proposed ontology fills this gap, being able to generate ontology scenarios for all corner case levels and execute them in simulation.

\section{Method}
\label{sec:method}

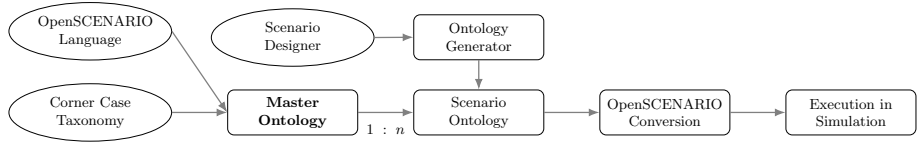
\begin{figure}[h!]
\noindent\resizebox{\textwidth}{!}{
\begin{tikzpicture}[
nodes={minimum height=3em, text width=7em, align=center},
node distance = 5mm and 12mm, 
block/.style = {draw, rounded corners, fill=#1,
                  minimum height=3em, text width=8em, align=center},
block/.default = white,
every edge/.append style = {draw=black!50, thick, -Latex}
                    ]
\node [draw, ellipse] (A) {Corner Case Taxonomy};
\node [draw, ellipse, above = of A] (B) {OpenSCENARIO Language};
\node [block, right = of A] (C) {\textbf{Master\\Ontology}};
\node [draw, ellipse, above = of C] (D) {Scenario Designer};
\node [block, right = of C] (E) {Scenario Ontology};
\node [block, above = of E,yshift=4pt] (H) {Ontology Generator};
\node [block, right = of E] (F) {OpenSCENARIO Conversion};
\node [block, right = of F] (G) {Execution in Simulation};

\draw[thick,anchor=west]   (A.east)  edge  (C.west);         
\draw[thick,anchor=west]   (B.east)  edge  (C.west);         
\draw[thick,anchor=west]   (C.east)  edge  node[below,yshift=4pt]{\small $1:n$}(E.west);  
\draw[thick,anchor=west]   (E.east)  edge  (F.west);   
\draw[thick,anchor=west]   (F.east)  edge  (G.west);  

\draw[thick,anchor=west]   (D.east)  edge  (H.west);   
\draw[thick,anchor=west]   (H.south)  edge  (E.north);

\end{tikzpicture}
}
\caption{Flow diagram of our proposed method for the description and generation of corner case scenarios. Based on a corner case taxonomy and the OpenSCENARIO language, a \textit{Master Ontology} was developed, containing all necessary attributes to describe complex scenarios. In a $1:n$ relation, ontologies describing individual scenarios can be derived. In an automated fashion, these scenarios are then converted into the OpenSCENARIO format, enabling the direct execution in simulation environments.}
\label{fig:main_approach}
\end{figure}

In order to generate corner case scenarios, we have developed a \textit{Master Ontology} which is the foundation for the creation of specific scenarios and provides the structure for all elements of a scenario. Based on this, all common corner case categories found in the literature can be addressed. For the creation of scenarios, we have developed an \textit{Ontology Generator} module, which is our interface to human \textit{Scenario Designers} which do not need any expertise in the field on ontologies in order to design scenarios. For each designed scenario, a concrete \textit{Scenario Ontology} is created. This is a major advantage over purely coded scenarios, as the complete scenario description is available in a human- and machine-readable form, which directly enables knowledge extraction, analysis, and further processing, such as exports into others formats or combinations of scenarios, for all created scenarios on any level of detail. Finally, our \textit{OpenSCENARIO Conversion} module converts this ontology into an OpenSCENARIO file, which can directly be simulated in the CARLA simulator. An overview can be found in Fig.~\ref{fig:main_approach}.

\vspace{3mm} 

\subsection{Master Ontology} \label{master_onto_sec}
At first, we describe the \textit{Master Ontology}, which is the skeleton of every concrete scenario. With its help, different scenarios can be described by instantiating the different classes, using individuals, and setting property assertions between them. The \textit{Master Ontology} is closely aligned to the OpenSCENARIO documentation \cite{OpenSCENARIO_main}, since the ontology is used for automatic generation of scenarios. Within the ontology, it is also possible to describe concrete categories of a corner cases, based on the categories introduced by Breitenstein et al.~\cite{breitenstein_corner_2021}.

\begin{figure}[ht]
\includegraphics[width = \textwidth]{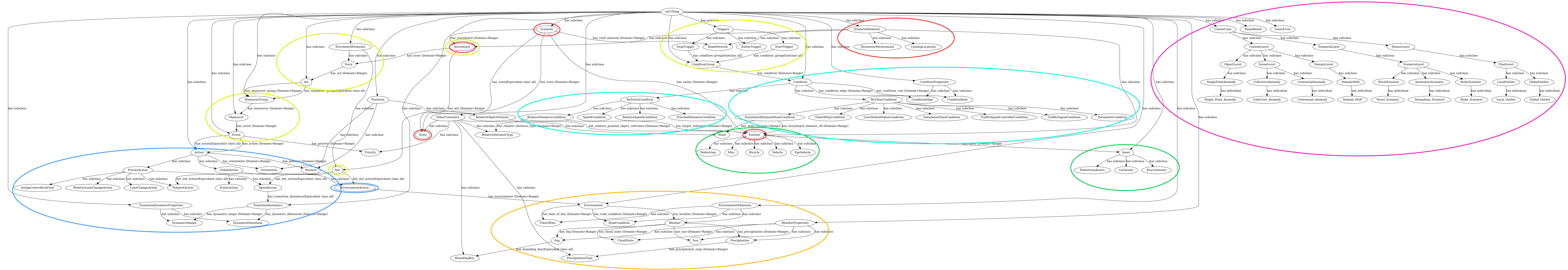}
\centering
\caption{Master ontology, best viewed at 1,600~\% and in color. The ontology is capable to describe scenarios based on the seven sections scenario and environment, entities, main scenario elements, actions, conditions, weather and time of day, and corner case level. These seven sections are further explained in Sec.~\ref{master_onto_sec}. Adapted from~\cite{Guneshka_Ontology_2022_BA}.}
\label{fig:master_ontology}
\end{figure}

The master ontology, as shown in Fig.~\ref{fig:master_ontology}, consists of 100 classes, 53 object properties, 44 data properties, 67 individuals, and 683 axioms. The 100 classes are either classes for the description of the corner case category or derived from the OpenSCENARIO documentation~\cite{OpenSCENARIO_main}, which means that the definitions of the different OpenSCENARIO elements can also be found there. They are used as parents for the different individuals we have created within the ontology. The 53 object properties and the 44 data properties are used to connect the different parts of a scenario, in order to embed individuals into concrete scenarios. For a better understanding and more structured explanation, the proposed \textit{Master Ontology} can be divided into seven main groups - Scenario and Environment, Entities, Main Scenario Elements, Actions, Conditions, Weather and Time, and Corner Case Level. We will describe these in more detail in the following, with each section marked with an individual color in Fig.~\ref{fig:master_ontology}. 

\textbf{Scenario and Environment (red).} In order to be able to describe a scenario, the \textit{Master Ontology} provides the Scenario class, which acts as the root of the ontology. Together with the scenario class, different object and data properties are provided. Those are used as connections between the different scenario elements, such as the Entities, Towns or the Storyboard. Towns are CARLA specific environments used in the ontology. CARLA allows users to create custom and thus arbitrary environments.

\textbf{Entities (green).} This group holdes the different entities Vehicle, Pedestrian, Bicycle, and Misc. For arbitrary Entities, the Misc class can be utilized. If specific movement patterns are wanted, the classes Vehicle, Pedestrian and Bicycle are also already available. The individuals can be then connected to 3D assets from the CARLA blueprint library~\cite{carla_assets}, which can be extended with external objects. This way, a \textit{Scenario Designer} is able to add arbitrary assets into a scenario.

\textbf{Main Scenario Elements (yellow).} The main scenario elements are used to build the core of any scenario. The highest level is the Storyboard, which includes an Init and a Story. A Story has at least one Act, which needs at least a StartTrigger and can optionally include a StopTrigger. Acts also are a container for different ManeuverGroups, that logically include Maneuvers. The Maneuvers then have to have minimum one Event, which is also activated by a StartTrigger. At last, each Event needs to include at least one Action. These are the main components of the OpenSCENARIO scenario description language and thus necessary parts of each scenario. For each of them, also a corresponding connecting property exists, i.e. \textit{has\_event, has\_action, has\_init\_action}. 

\textbf{Actions (dark blue).} To be able to describe the maneuvers of the different Entities, different Actions are represented within the Ontology. Those include, e.g., TeleportAction, which sets the position of an Entity, or RelativeLaneChangeAction, which describes a lane change of an Entity. 

\textbf{Conditions (light blue).} 
As part of the StartTrigger and StopTrigger elements, Conditions are used to activate them. Conditions are divided into the two subclasses ByEntityCondition and ByValueCondition. In general the difference between those two is that the ByEntityCondition is always in regard to an entity, i.e., how close a vehicle is to another vehicle, while the ByValueCondition is always in regard to a value, i.e., the passed simulation time. Depending on the type of the Condition, different values must be met in order for the StartTrigger or StopTrigger to get activated. As an example, the SimulationTimeCondition can be used as a trigger with respect to the simulation time, using arithmetic rules.

\textbf{Weather and Time of Day (orange).} To set the weather, the underlying CARLA town can be modified individually. This includes the weather conditions, which are subdivided into fog, precipitation, and the position of the sun. Also, the time of day can be set. 

\begin{figure}[h!]
    \centering
    \includegraphics[width=1.0\textwidth]{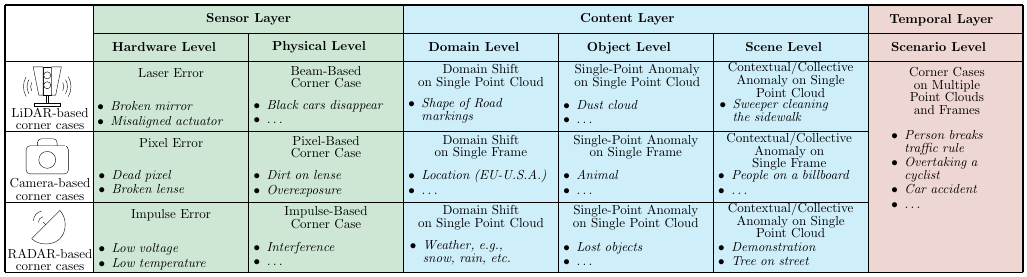}
    \caption{Corner Case Categorization from Heidecker et al.~\cite{Heidecker_2021}. The columns show different layers and levels of corner cases, while the rows are related to specific sensors. For each combination, examples are provided.}
    \label{fig:categorization corner cases}
\end{figure}

\textbf{Corner Case Level (pink).} In the long tail of rare scenarios, each can be related to a specific corner case level. The first ones to propose an extensive taxonomy on these were Breitenstein et al.~\cite{corner_case_first} with a focus on camera-based sensor setups. This taxonomy was extended by Heidecker et al.~\cite{Heidecker_2021} to generalize it to a set of three top-level layers, namely Sensor Layer, Content Layer, and Temporal Layer, as shown in Fig.~\ref{fig:categorization corner cases}. In this work, we focus on camera-related corner cases, which is why the master ontology uses a mixed model, where the top level layers from Heidecker et al. and the underlying corner case levels from Breitenstein et al. are used, as shown in Fig.~\ref{fig:master_ontology}, making a future extension of the master ontology to further sensors effortless, as they fall into the same top level layers. Occurrences on the hardware or physical-level, such as dead pixels or overexposure, can be simulated with subsequent scripts during the simulation phase. Details on those corner cases can be placed in the individual scenario ontologies by creating specific individuals of the respective corner case classes of the \textit{Master Ontology}.

Next to those groups, additional 67 individuals exist, which are divided into Constants and Default Individuals. There are two types of constants: OpenSCENARIO Constants, such as arithmetic or priority rules, and CARLA Constants, such as assets. The default individuals are used to help a \textit{Scenario Designer} to create scenarios faster and easier. These include common patterns, such as default weather conditions or a trigger, which activates when the simulation starts running.
In addition, a default ego vehicle is also included in the \textit{Master Ontology}, which has a set of cameras and a BoundingBox attached to it. As the last part of the ontology, the 683 axioms represent the connections and rules between the entities and the properties within the ontology, along with the individuals. 

\subsection{Scenario Ontology Generation}
The manual creation of ontologies is a very time-consuming, exhausting and error-prone process, which additionally needs expertise in the general field of ontologies and related software. Thus, and to ensure, that the \textit{OpenSCENARIO Conversion} module functions properly, we have developed the \textit{Ontology Generator} module, which takes as input a scripted version of a scenario and creates a \textit{Scenario Ontology} as a result. The concept behind the \textit{Ontology Generator} is to use the \textit{Master Ontology} as a base for a scenario description and automatically create the necessary individuals and property assertions between them. To read and write ontologies, we utilize the python library Owlready2~\cite{owl_ready}. The \textit{Master Ontology} is read by the \textit{Ontology Generator}, and it uses, depending on the scenario, all classes, properties, and default individuals needed. The result is a new \textit{Scenario Ontology}, which has the same structure as the \textit{Master Ontology} with respect to classes and properties, but includes newly created individuals for the scenario designed by the \textit{Scenario Designer}.  

\begin{table}[h]
\resizebox{\textwidth}{!}{%
\begin{tabular}{@{}cccc@{}}
\toprule
Main Scenario Elements                & Entities             & Actions                       & Environment and Weather \\ \midrule
newScenario                  & newEgoVehicle        & newEnvironmentAction          & newEnvironment          \\
newStoryboard                & newCar               & newSpeedAction                & newTimeOfDay            \\
newInit                      & newPedestrian        & newTeleportActionWithPosition & newWeather              \\
newStory                     & newMisc              & newTeleportActionWithRPAO     & newFog                  \\
newAct                       &                      & newRelativeLaneChangeAction   & newPrecipitation        \\
newManeuverGroup             &                      &                               & newSun                  \\
newManeuver                  &                      &                               & changeWeather                        \\
newEvent                     &                      &                               &                         \\
newAction                    &                      &                               &                         \\
\multicolumn{1}{l}{}         & \multicolumn{1}{l}{} & \multicolumn{1}{l}{}          & \multicolumn{1}{l}{}    \\ \midrule
Conditions                   & Assets               & Other                         &                         \\ \midrule
newSimulationCondition       & newAsset             & newStopTrigger                &                         \\
newRelativeDistanceCondition & getBicycleAssets     & newStartTrigger               &                         \\
newStoryboardESCondition     & getCarAssets         & newRoadCondition              &                         \\
newTraveledDistanceCondition & getPedestrianAssets  & newTransitionDynamics         &                         \\
                             & getMiscAssets        & setCornerCase                              &                        
\end{tabular}%
}
\vspace{3mm} 
\caption{Overview of methods of the \textit{Ontology Generator} module, which are available to a human \textit{Scenario Designer}. The methods are divided in seven groups, five from which were introduced in Sec.~\ref{master_onto_sec}. The groups Assets and Other are related to 3D objects and miscellaneous functions, respectively.}
\label{tab:ontogen_methods1}
\end{table}

\SetKwComment{Comment}{//}{}
\begin{algorithm}[h!]
\caption{Creation of a scenario ontology with the Ontology Generator, where the ego vehicle enters a foggy area (incl. abstract elements) }\label{algo:onto_generator}
$import~OntologyGenerator~as~OG$

$import~MasterOntology~as~MO$

$ego\_vehicle \gets MO.ego\_vehicle\;$ \Comment{Default ego vehicle individual}

$weather\_def \gets MO.def\_weather\;$
\Comment{Default weather individual}

$Initialize~teleport\_action(ego\_vehicle), speed\_action(ego\_vehicle)$

$ init\_scenario \gets OG.newInit(speed\_action,teleport\_action,weather\_def)$

\Comment{Starting conditions for storyboard}

$Initialize~traveled\_distance\_condition$

$ Trigger\gets OG.newStartTrigger(traveled\_distance\_condition)$

\Comment{Trigger Condition: Ego vehicle travelled defined distance}

$Initialize~weather(sun,fog,precipitation)$

$Initialize~time\_of\_day, road\_condition$

$env \gets OG.newEnv(time\_of\_day, weather, road\_condition)$

$env\_action \gets OG.newEnvAction(env)$

\Comment{Foggy environment after trigger}

$Initialize~Event,Maneuver,ManeuverGroup,Act,Story,Storyboard$

\Comment{Necessary OpenSCENARIO elements}
$Export~ScenarioOntology$
\end{algorithm}

Since the \textit{Master Ontology} is built based on the OpenSCENARIO documentation~\cite{asam_openscenario}, which is a very powerful and flexible framework, it allows for many possible combinations. This gives a \textit{Scenario Designer} a large flexibility with respect to the design of new scenarios. The \textit{Ontology Generator} is well documented. This way, no prior experience with the OpenSCENARIO format is necessary. Table \ref{tab:ontogen_methods1} provides an overview of the functions available to the human \textit{Scenario Designer} to create scenarios.

With the help of the \textit{Ontology Generator}, every part which was defined within the \textit{Master Ontology} can be utilized. For example, the functions within the first group shown in Table~\ref{tab:ontogen_methods1}, are used to create every scenario main element. Algorithm~\ref{algo:onto_generator} shows, how a partly abstracted implementation, as done by a \textit{Scenario Designer}, looks like. The full example can be found in the GitHub repository. In Sec.~\ref{sec:evaluation} we demonstrate an exemplary \textit{Scenario Ontology} which was generated by the \textit{Ontology Generator}. In this demonstration, the scenario ontology from Algorithm~\ref{algo:onto_generator} is related to the visualization in Fig.~\ref{fig:fog_ontology}. 

\subsection{Scenario Simulation}
After a scenario is described with the help of individuals within a scenario ontology, it is being read by the \textit{OpenSCENARIO Conversion} module, as shown in Fig.~\ref{fig:main_approach}. From these concrete scenarios, the conversion module generates OpenSCENARIO files. These can can be directly simulated without any further adjustments. Since the OpenSCENARIO files include simulator-specific details, we have focused on the CARLA\footnote{CARLA version 0.9.13 was utilized} simulation environment~\cite{carla}. In an earlier stage, we were also able to show compatibility with the esmini~\cite{esmini} environment.

When the \textit{Ontology Generator} module is used to create the scenario ontologies, their structural integrity is ensured, which is a necessary requirement for the conversion module. This means that each scenario ontology is correctly provided to be processed by the conversion module. Theoretically, scenario ontologies could also be created manually to be processed by the conversion module. However, human errors are likely, preventing the correct processing by the conversion module.

While each scenario ontology is able to cover multiple corner cases, the created ontologies are fully modular. This means, given the same environment, our method is capable of combining multiple, already existing scenario ontologies into a new single scenario ontology. In such cases, where the number of scenario individuals is $n>1$, a pre-processing stage is triggered which extends the ontology to combine all $n$ provided scenarios into a single new scenario $S_{fusion}$. For this purpose, this stage creates a new scenario, storyboard, and init. Subsequently, for every included scenario, the algorithm goes through its stories, entities, and init actions and merges them in $S_{fusion}$. For the final creation of the OpenSCENARIO file, the conversion module utilizes the property assertions between individuals to create the according python objects, which are then used by the PYOSCX library~\cite{scenario_generation} to create the OpenSCENARIO file. These files can then be read by the ScenarioRunner~\cite{scenario_runner} and executed in CARLA. In the following Sec.~\ref{sec:evaluation}, we demonstrate a set of ten simulated scenarios.

\section{Evaluation}
\label{sec:evaluation}

For the evaluation, we have created a diverse scenario catalog containing scenarios from all corner case levels. Following Breitenstein et al.~\cite{Breitenstein_2020}, these cover different levels of complexity and thus criticalities, starting with simpler sensor layer cases and ending with highly complex temporal layer corner cases. For the qualitative evaluation we show the feasibility of the approach as shown in Fig.~\ref{fig:main_approach} and demonstrate it with a set of ten scenarios, that descriptions made by a human \textit{Scenario Designer} get translated into proper \textit{Scenario Ontologies} and correctly simulated.

\begin{table}[h!]
\centering
\resizebox{\columnwidth}{!}{%
\begin{tabular}{@{}llcr@{}}
\toprule
\# & Corner Case Level             & Individuals & Scenario Description \\ \midrule
(a)  & Sensor Layer - Hardware Level & -         & Dead Pixel: Camera sensor affected           \\
(b)  & Content Layer - Domain Level  & 94           & Domain Shift: Sudden weather change         \\
(c)  & Content Layer - Object Level  & 93           & Single-Point Anomaly: Unknown object on the road \\
(d)  & Content Layer - Scene Level   & 164           & Collective Anomaly: Multiple known objects on the road  \\
(e)  & Content Layer - Scene Level   & 111           & Contextual Anomaly: Known non-road object on the road    \\
(f)  & Temporal Layer - Scenario Level                & 94           & Novel Scenario: Unexpected event in another lane\\
(g)  & Temporal Layer - Scenario Level                & 104           & Risky Scenario: A risky maneuver
\\
(h)  & Temporal Layer - Scenario Level            & 95           & Anomalous Scenario: Unexpected traffic participant behaviour\\
(i)  & Combined: (d) and (f)                    & 156           & Combined: Collective and Novel Scenario            \\
(j) & Combined: (f) and (h)                     & 122           & Combined: Novel and Anomalous Scenario          \\ \bottomrule
\end{tabular}%
}
\vspace{3mm} 
\caption{Overview of scenario ontologies, which were derived from the master ontology and subsequently executed in simulation. These exemplary scenarios cover all corner case categories.}
\label{tab: examples}
\end{table}

For the selection of the exemplary corner case scenarios, we considered three types of sources. First, we used examples provided by the literature, such as the ones provided by Breitenstein et al.~\cite{breitenstein_corner_2021}. Second, various video sources, such as third-person videos, and dash-cam videos of traffic situations~\cite{youtube_dash} were used for inspiration. Third, multiple brainstorming sessions took place, where personal experiences were collected. Afterwards, we narrowed the selection down to a set of eight representative scenarios. Two more were created by combining two of those eight scenarios. An overview of these scenarios can be found in Table~\ref{tab: examples}. Visualizations of all scenarios can be found in Fig. \ref{fig:examples}.

\vspace{3mm} 
In the \textit{a) Dead Pixel} scenario, an arbitrary scenario can be chosen, where only the sensor itself will be affected. The \textit{b) Domain Shift} is a sudden weather change, where we created a scenario where the ego vehicle suddenly drives into a dense fog. For the \textit{c) Single-Point Anomaly}, we chose to simulate a falling vending machine on the road. This scenario also inspired the next scenario, \textit{d) Contextual Anomaly}, which also has falling objects on the road, but in this case the objects are traffic signs. This can be considered, for example, in a very windy environment. As a \textit{e) Collective Anomaly}, we chose to simulate a lot of running pedestrians in front of the ego vehicle, which can for example happen during a sports event. For the \textit{f) Novel Scenario}, we described a scenario where a cyclist performs unexpected maneuvers in the opposite lane. The \textit{g) Risky Scenario} which we chose is a close cut-in maneuver in front of the ego vehicle. The last corner case category is the \textit{h) Anomalous scenario}, for which we have chosen a pedestrian who suddenly runs in front of the ego vehicle. To demonstrate the scalability of our approach, we have also combined scenarios. In the \textit{i)} combination of \textit{Novel Scenario} and the \textit{Collective Scenario}, a lot of running pedestrians are in front of the ego vehicle, while a cyclist performs unexpected maneuvers in the opposite lane, next to the pedestrians. In addition, we also merged the \textit{Novel Scenario} with the \textit{Anomalous Scenario}, resulting in \textit{j)} a pedestrian walking in front of the ego vehicle and the cyclist.

\afterpage{ 
\begin{figure*}[ht!]
\setlength\tabcolsep{6pt} 
\centering
\begin{tabular}{@{} r M{0.15\linewidth} M{0.15\linewidth} M{0.15\linewidth} M{0.15\linewidth} @{}} %
& $t_0$ & $t_1$ & $t_2$\\
    \begin{subfigure}{0.2\linewidth}\caption{Dead Pixel}
    \label{subfig:a} 
    \end{subfigure} 
    & \includegraphics[width=\hsize]{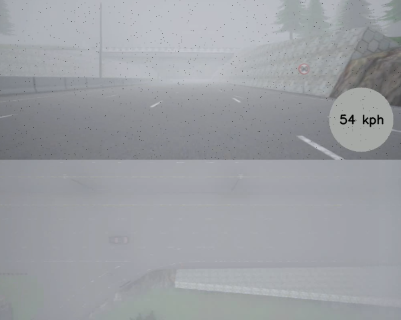}
    & \includegraphics[width=\hsize]{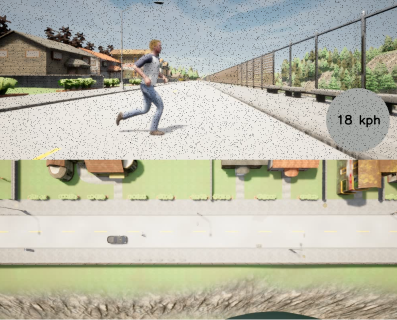}
    & \includegraphics[width=\hsize]{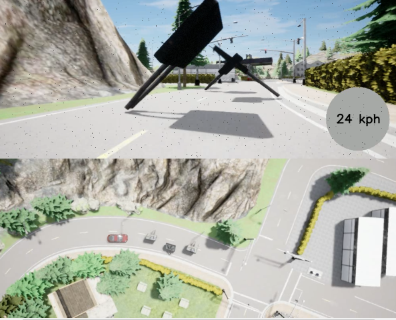}\\
    \addlinespace
    \begin{subfigure}{0.22\linewidth}\caption{Domain Shift}
    \label{subfig:b} 
    \end{subfigure} 
    & \includegraphics[width=\hsize]{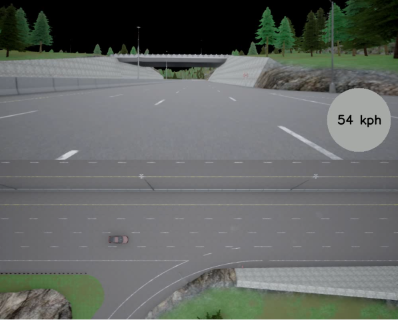} 
    & \includegraphics[width=\hsize]{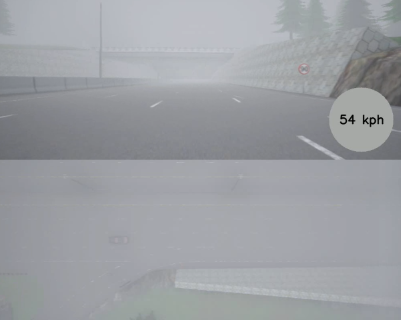}
    & \includegraphics[width=\hsize]{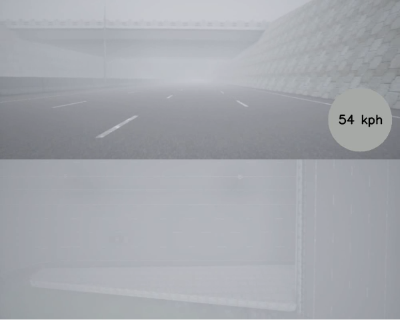}\\ 
    \addlinespace
    \begin{subfigure}{0.31\linewidth}\caption{Single-Point Anomaly}
    \label{subfig:c} 
    \end{subfigure} 
    & \includegraphics[width=\hsize]{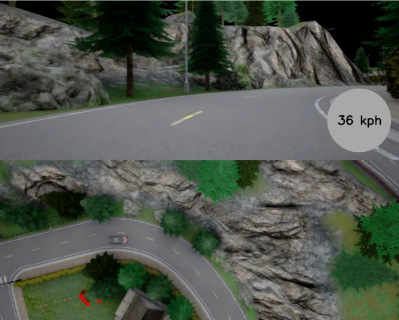}
    & \includegraphics[width=\hsize]{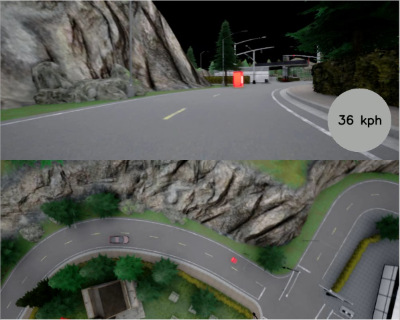}
    & \includegraphics[width=\hsize]{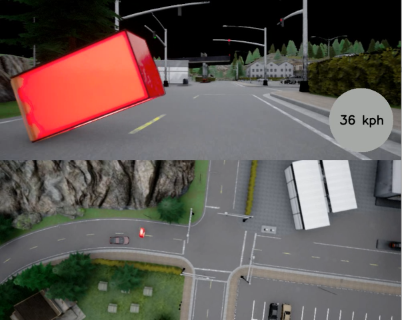}\\
    \addlinespace
    \begin{subfigure}{0.28\linewidth}\caption{Collective Anomaly}
    \label{subfig:d} 
    \end{subfigure} 
    & \includegraphics[width=\hsize]{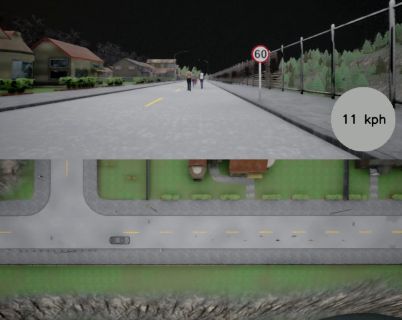}
    & \includegraphics[width=\hsize]{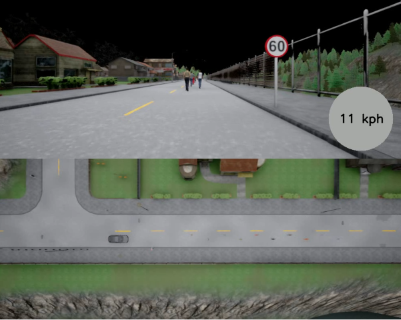}
    & \includegraphics[width=\hsize]{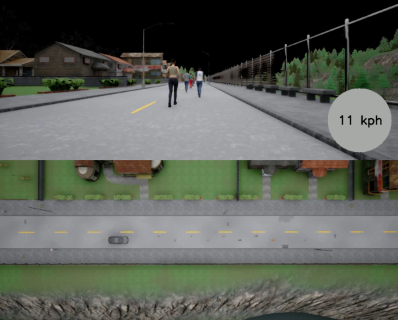}\\
    \addlinespace
     \begin{subfigure}{0.29\linewidth}\caption{Contextual Anomaly}
    \label{subfig:e} 
    \end{subfigure} 
    & \includegraphics[width=\hsize]{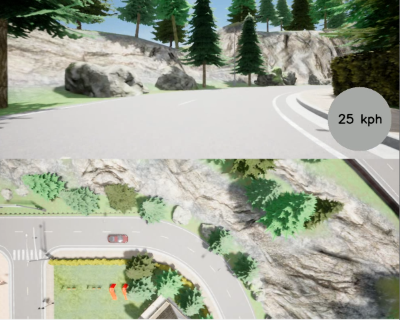}
    & \includegraphics[width=\hsize]{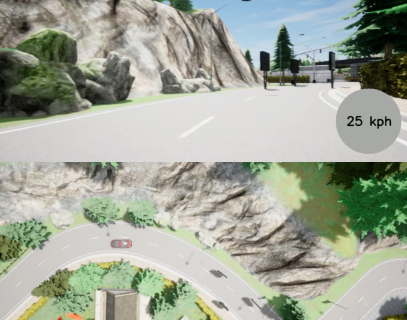}
    & \includegraphics[width=\hsize]{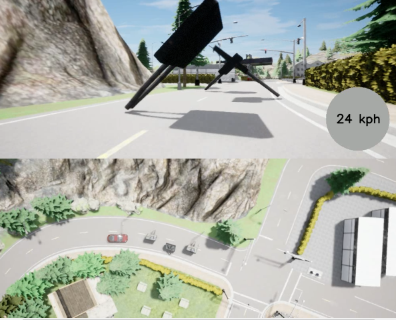}\\
    \addlinespace
    \begin{subfigure}{0.23\linewidth}\caption{Novel Scenario} 
    \label{subfig:f} 
    \end{subfigure} 
    & \includegraphics[width=\hsize]{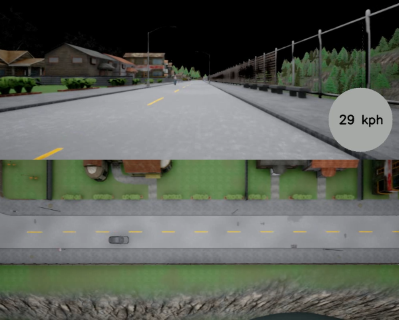}
    & \includegraphics[width=\hsize]{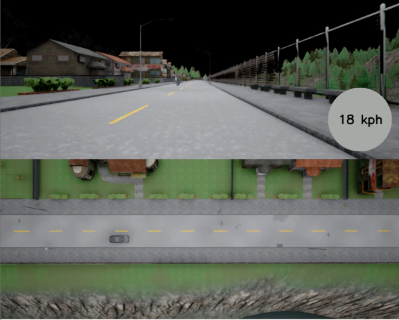}
    & \includegraphics[width=\hsize]{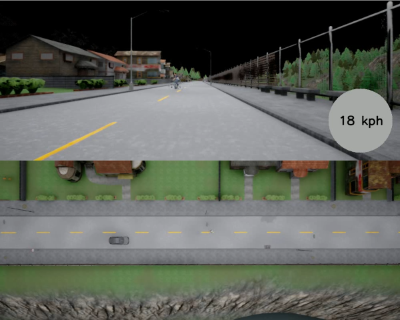}\\
    \addlinespace
        \begin{subfigure}{0.23\linewidth}\caption{Risky Scenario} 
    \label{subfig:g} 
    \end{subfigure} 
    & \includegraphics[width=\hsize]{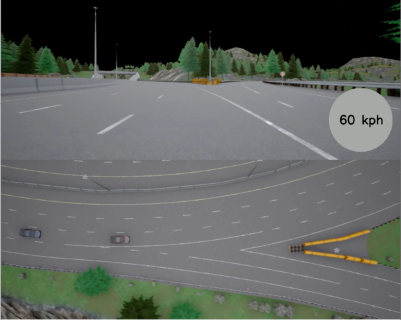}
    & \includegraphics[width=\hsize]{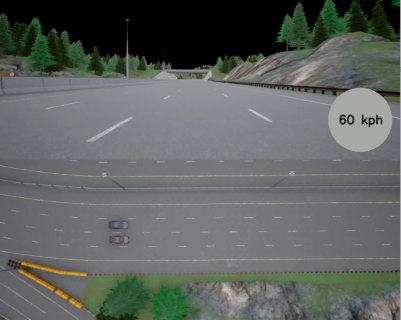}
    & \includegraphics[width=\hsize]{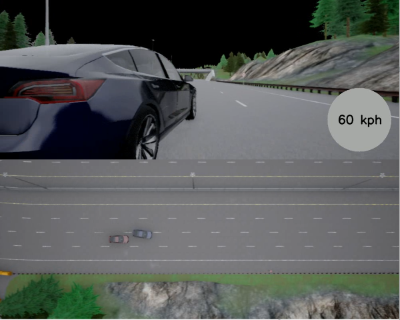}\\
    \addlinespace
    \begin{subfigure}{0.29\linewidth}\caption{Anomalous Scenario} 
    \label{subfig:h} 
    \end{subfigure} 
    & \includegraphics[width=\hsize]{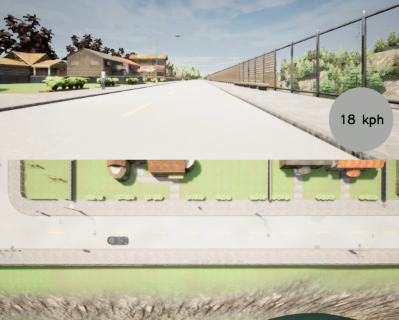}
    & \includegraphics[width=\hsize]{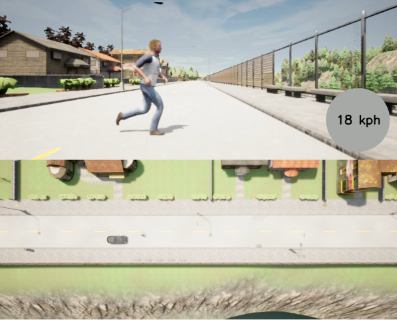}
    & \includegraphics[width=\hsize]{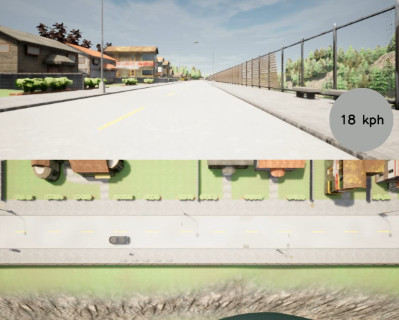}\\
    \addlinespace
    \begin{subfigure}{0.2\linewidth}\caption{Merged (d)(f)} 
    \label{subfig:i} 
    \end{subfigure} 
    & \includegraphics[width=\hsize]{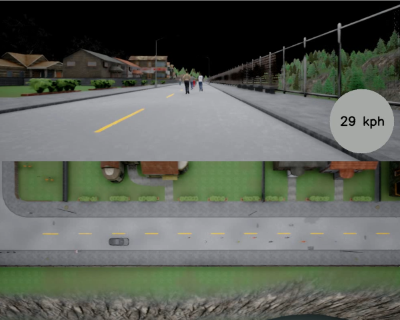}
    & \includegraphics[width=\hsize]{images/framescc/merge1.PNG}
    & \includegraphics[width=\hsize]{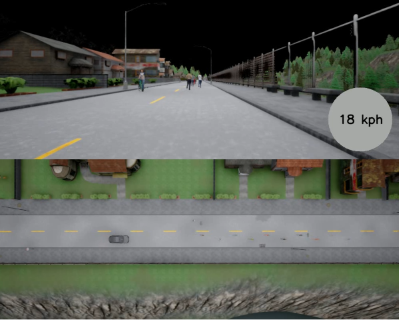}\\
    \addlinespace
    \begin{subfigure}{0.2\linewidth}\caption{Merged (f)(h)} 
    \label{subfig:j} 
    \end{subfigure} 
    & \includegraphics[width=\hsize]{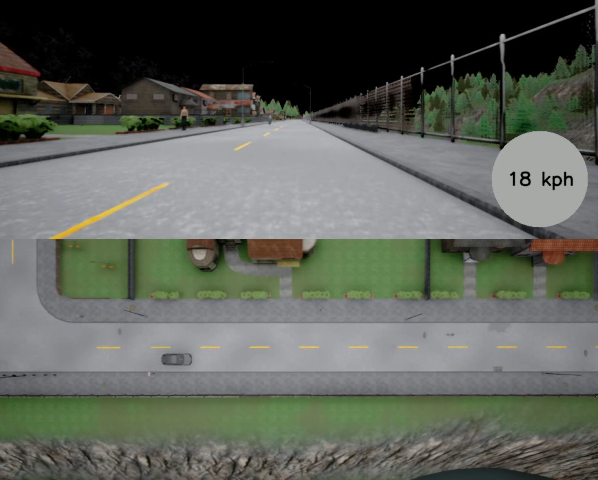}
    & \includegraphics[width=\hsize]{images/framescc/merge_pedestrian_bicycle1.PNG}
    & \includegraphics[width=\hsize]{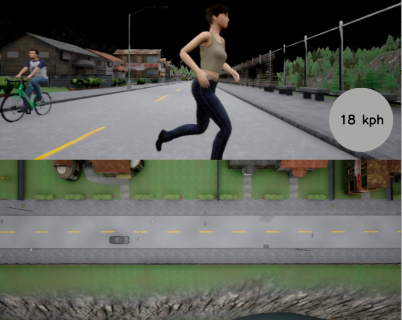}\\
    \addlinespace
\end{tabular}
\caption{Visualization of the realized corner case scenarios as listed in Table~\ref{tab: examples}.}
\label{fig:examples}
\end{figure*}
\clearpage
}

\subsection{Scenario Ontologies}
\begin{figure}[h]
    \centering
    \includegraphics[width = \textwidth]{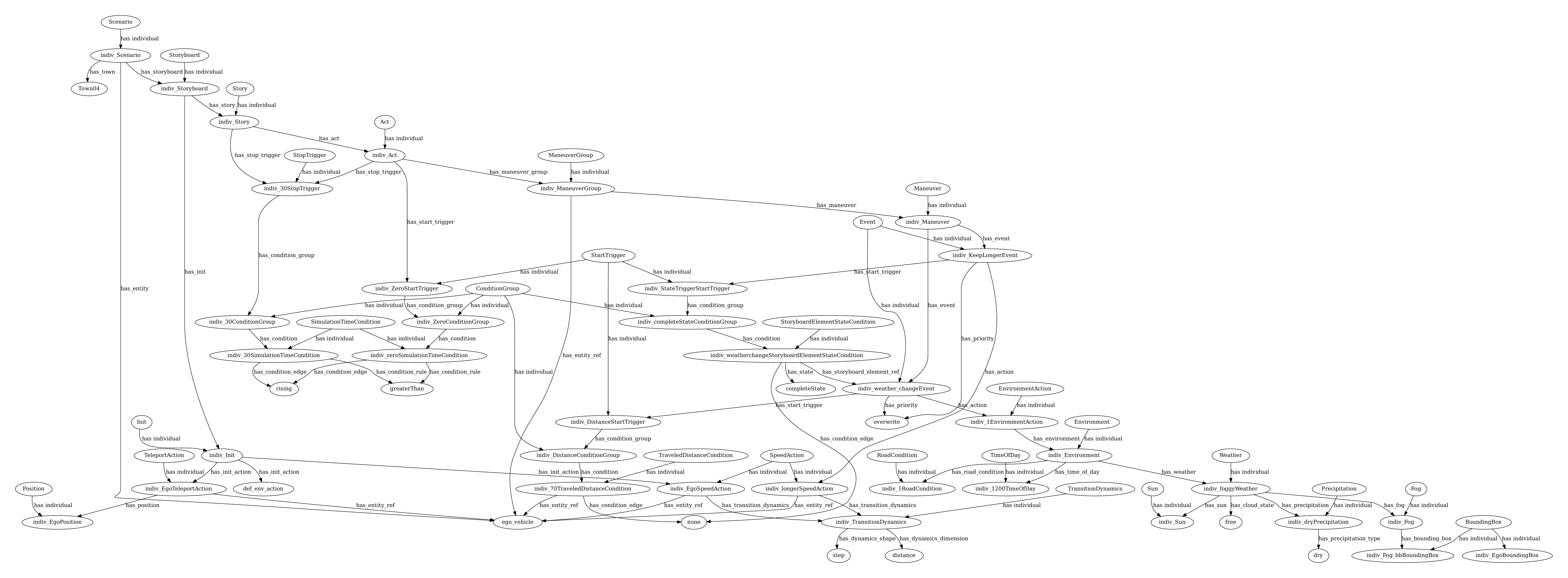}
    \caption{Scenario ontology describing a vehicle entering a foggy area. Best viewed at 1,600~\%. The ontology describes the scenario based on the seven sections scenario and environment, entities, main scenario elements, actions, conditions, weather and time of day, and corner case level. Reprinted from~\cite{Guneshka_Ontology_2022_BA}.}
    \label{fig:fog_ontology}
\end{figure}

At the core of each demonstrated scenario lies a \textit{Scenario Ontology}. In the following, we present the construction of an exemplary scenario and describe how it is represented in the \textit{Scenario Ontology} with individuals. An example graph of the exemplary ontology can be seen in Fig.~\ref{fig:fog_ontology}. The ontology has 94 individuals, which means that 27 new individuals were created, since the \textit{Master Ontology} has 67 default individuals.

In Fig.~\ref{fig:fog_ontology}, every used class, property and individual is presented. Each 
individual, which name starts with "\textit{indiv\_}", is a newly created part of the \textit{Scenario ontology}, every other individual is either a default or a constant.
The graph starts from the top with the \textit{Scenario} individual, which is connected to a CARLA town and a newly created Storyboard. As mentioned earlier, every \textit{Storyboard} has an \textit{Init} and a \textit{Story}. In this particular \textit{Init}, there are only the \textit{Action}s, which are responsible for the position and the speed of the \textit{EgoVehicle}, and connections to the default \textit{EnvironmentAction}. The most interesting part of this \textit{Scenario} however can be found deep within the \textit{Story} - namely the second \textit{EnvironmentAction}, which creates a dense fog inside the scenario. This \textit{Action} gets triggered by the \textit{indiv\_DistanceStartTrigger}, which has a \textit{TraveledDistanceCondition} as a \textit{Condition}. Since this type of \textit{Condition} is an \textit{EntityCondition}, it requires a connection to an \textit{Entity}, in our case the \textit{ego\_vehicle}. This \textit{StartTrigger} gets activated when the \textit{ego\_vehicle} has travelled a certain distance. After this \textit{Event} is executed, the Scenario comes to its end.

Descriptions and visualizations of the remaining nine \textit{Scenario Ontologies} can be found in \cite{Guneshka_Ontology_2022_BA}, and the code for all ten scenarios, in order to recreate them, can be found in the GitHub repository.

\section{Conclusion}
\label{sec:conclusion}
Our work focuses on the generation of rare corner case scenarios in order to provide additional input training data as well as test and evaluation scenarios. This way, it contributes to potentially safer deep neural networks, which might be able to become more robust to anomalies. The proposed \textit{Master Ontology} is the core of our approach and enables the creation of specific scenarios for all of the corner case levels developed by Breitenstein et al.~\cite{breitenstein_corner_2021}. From the one \textit{Master Ontology}, concrete scenario ontologies can be derived. The whole process after the design of such an ontology is fully automated, which includes the generation of ontologies itself, based on the human input for the \textit{Ontology Generator} as well as the conversion into the OpenSCENARIO format. This allows for a direct simulation of the scenarios in the CARLA simulation environment without any further adjustments. Since ontologies are a highly complex domain, we have put an emphasis on the design of the \textit{Ontology Generator} module, which does not require expertise in the field of ontologies for a human to create scenarios. We have demonstrated our approach with a set of ten concrete scenarios, which cover all corner case levels. 

\vspace{3mm} 
As an outlook, we would first like to discuss existing limitations of our work. At the moment, the master ontology is focused on camera-related corner cases, which is why the current implementation only includes hardware-related corner cases for camera sensors. Also, the subsequent scripts for camera-related corner cases are currently not triggered automatically, but manually. For the \textit{Master Ontology}, we have implemented a vast body of the OpenSCENARIO standard in order to demonstrate our designed scenarios. However, for future scenarios, modifications of the master ontology might be necessary. We have provided instructions for such extensions in our GitHub repository. While the master ontology supports arbitrary objects, in our demonstrated scenarios we were only able to utilize assets which were already included within CARLA, as we ran into compilation issues with respect to the Unreal Engine and the utilized CARLA version. Our learnings as well as ideas to address this issue can also be found in the GitHub repository.

\vspace{3mm} 
Finally, we would like to point out future directions. The extraction of corner case levels from the ontology itself could be automated, e.g., with knowledge extraction methods based on Semantic Web Rule Language (SWRL) rules. However, this is a challenging field itself. Based on the generated scenarios, which were created by human \textit{Scenario Designers}, automated variations can be introduced to drastically increase the number of available scenarios, as shown by~\cite{Li_TestGeneration_2020,Hermann_Using_2022_DATE,functional_to_logical}. This way, a powerful combination of knowledge- and data-driven scenario generation can be achieved for the long tail of rare corner cases.
\section*{Acknowledgment}
\label{sec:ack}
This work results partly from the project KI Data Tooling (19A20001J) funded by the  Federal Ministry for Economic Affairs and Climate Action (BMWK).


\clearpage
%
%
\bibliographystyle{splncs04}
\bibliography{egbib}
\end{document}